\documentclass[%
 reprint,
 amsmath,amssymb,
 aps,
]{revtex4-1}

\usepackage{graphicx}
\usepackage{dcolumn}
\usepackage{bm}

\usepackage{algorithm}
\usepackage{algorithmic}

\begin{document}

\preprint{APS/123-QED}

\title{Constants of Motion for Conserved and Non-conserved Dynamics}

\author{Michael F. Zimmer}
\email{zim@neomath.com}

\date{\today}

\begin{abstract}
This paper begins with a dynamical model that was obtained by applying a machine learning technique (FJet) to time-series data;
this dynamical model is then analyzed with Lie symmetry techniques to obtain constants of motion.
This analysis is performed on both the conserved and non-conserved cases of the 1D and 2D harmonic oscillators.
For the 1D oscillator, constants are found in the cases where the system is underdamped, overdamped, and critically damped.
The novel existence of such a constant for a non-conserved model is interpreted as a manifestation of the conservation of energy of the {\em total} system (i.e., oscillator plus dissipative environment). 
For the 2D oscillator, constants are found for the isotropic and anisotropic cases, including when the frequencies are incommensurate; it is also generalized to arbitrary dimensions.
In addition, a constant is identified which generalizes angular momentum for all ratios of the frequencies.
The approach presented here can produce {\em multiple} constants of motion from a {\em single}, generic data set.
\end{abstract}

\maketitle



\section{Introduction}
\label{sec:intro}

As discussed in \citep{MFZ-FJet-arxiv-current}, determining a constant of motion from time series data is part of a larger program
of using machine learning to discover physical laws, in particular, differential equations (DEs).  In that paper, the key points were identified as
\begin{enumerate}
\item Extrapolate model beyond training data times.
\item Determine underlying DE.
\item Determine parameter dependencies in DE.
\item Estimate stability/accuracy of model.
\item Determine related domain knowledge. 
\end{enumerate}
This paper seeks to address item \#5, that of computing domain knowledge (e.g., constants of motion) using that time series data.
The first four items were already addressed in \citep{MFZ-FJet-arxiv-current}.

The approach introduced herein assumes the results of \citep{MFZ-FJet-arxiv-current} as a starting point.  
That is, it assumes a model has already been found for a 1D harmonic oscillator (via the FJet method), in both the undamped and damped cases.
Thus, the previous paper was largely a machine learning (ML) effort, while this paper will be mainly analytical.

The new approach introduced herein relates the small/infinitesimal changes that are modeled by FJet,
with the small changes that are studied by Lie symmetry techniques for DEs.  
Since such techniques were in part designed to compute constants of motion, it forms a natural synergy with FJet for this investigation.
Also, notice that both of these techniques can be used when dissipation is present,
thus allowing the computation of a constant for both conservative {\em and} dissipative dynamics.
This versatility in the presence of dissipation offers it an advantage compared to methods by other authors.

\subsection*{Previous Work}
\label{sec:previous}

The focus here is mainly on work that has been done to identify symmetries/invariants of a physical system using ML techniques.
Where appropriate, it will again be discussed in the Comparison section (Sec.~\ref{sec:comparison}).

{\em \underline{Autoencoders (AE)}}:
In this case, the idea is to use an AE on a time-series to obtain a compressed representation in the latent variables of an AE, 
and then impose a physical prior on them.  (The reader may recall that an AE is a neural network (NN) whose multiple layers first encode and then decode the input.  It contains a middle layer with reduced capacity that forces a compressed representation there.)
This approach was taken by \citep{Greydanus-HNN-2019}, where they imposed Hamilton's equations on the latent variables, and were then able to study the energy.
Closely related to this, an AE was also used to enforce the Euler-Lagrange equations \citep{Cranmer-LNN-2020} on the latent space.

In an application to QCD and spin glasses, the authors of \citep{ADecelle-2019} used an AE trained with gauge-related input and outputs to
force the latent variables to represent gauge orbits for the $Z_2$ gauge symmetry.

{\em \underline{Embedding Layer}}:
In \citep{SWetzel-2020} the authors used Siamese NNs on grouped data to discover invariants for several physical theories.
Specifically, the data was created in pairs, so that some of the data shared an invariant, while the remainder did not.
Next, they modeled the embedding layer (i.e., the next to last layer) with polynomial regression to find a functional representation of an invariant.
Proceeding in this way, they recovered the space-time interval invariant for Lorentz boosts, 
the invariants ${\bf E}  \cdot {\bf B}$ and ${\bf B}^2-{\bf E}^2$ for electric ({\bf E}) 
and magnetic ({\bf B}) fields under a Lorentz boost, and the angular momentum and energy for motion in a central field.

In \citep{SKrippendorf-2021} the authors addressed the problem of classification of consistent vacua in string theory,
with a focus on complete intersection Calabi-Yau manifolds (CICYs).
Their target data were the topological invariant Hodge numbers $h^{1,1}$, $h^{1,2}$,
while the input data was formed from features of a graphical representation of configuration matrix parameterizations of the CICYs.
They identified clusters in the embedding layer corresponding to particular symmetries, and used a regression analysis (with PCA) to determine the Lie generators for those symmetries.
In addition, they examined several related test problems, where they used tSNE to identify clusters in the embedding layer.

{\em \underline{Specialized Loss Function}}:
In \citep{SHa-2021} the authors identified invariants from grouped data, 
where each group was expected to have its own value for a known invariant.  
They used a loss function which reduced the intra-group variance, while simultaneously increasing the inter-group variance.
Their examples included synthetic data for three test problems: the Lotka-Volterra model, a Kepler problem, and experimental data for a double pendulum.
The functional form of the invariant for one of the test problems was recovered using polynomial regression.

{\em \underline{Manifold Learning}}:  
In \citep{YMototake-2021} the author pursued an approach based on preserving the invariance of a (conservative) Hamiltonian 
under a continuous symmetry (i.e., time evolution).
From this, the author was able to determine an invariant for some conservative model systems.

In \citep{ZLiu-2021-manifold} the authors used manifold embedding techniques \citep{SSaremi-2019} to study the dynamics of
several systems: 1D oscillator, double pendulum, magnetic mirror, two planetary motion examples.
For these examples they determined the expected number of invariants and their functional form (using a symbolic regression dictionary ).
In addition, they applied it to a planetary system that has dissipation (i.e., a decaying orbit owing to tidal forces),
and showed how they can identify conserved quantities for a limited amount of time.

{\em \underline{Analytical}}:
There have been analytical investigations that don't use ML that are nevertheless relevant to the work presented here.
The symmetry group for the 2D anisotropic harmonic oscillator was determined by \citep{VDulock-1965} using a Hopf mapping.
In \citep{ASinha-2023,ASinha-2023b}, the authors computed an invariant for in the anisotropic case, using a technique that applies to Hamiltonian systems.
Lie symmetry techniques have been applied to the damped harmonic oscillator by \citep{JCervero-1984,CMuriel-2009,UDas-2013}
although from different vantage points than the present paper.
Also, Lie symmetry techniques in general \citep{Olver-1993,Stephani-1989} are a central theme in this paper.

{\em \underline{Adjacent Work}}:
In \citep{JYang-2023} the authors used an AE to force the latent variables (middle layer) to embody a GL symmetry,
which was related to symmetries expressed in the original variable vis-\`a-vis the encoder/decoder.
Their research has not yet taken the next step to identify invariants in the original variables.
Also, besides this particular contribution, there is other work on incorporating Lie symmetries into NNs,
but space does not permit a more thorough review.

A bourgeoning field of research has been to incorporate symmetries such as translation / rotation invariance into the architecture of NNs \citep{TCohen-2016}.
If such NNs are to describe the dynamics of physical systems, this is certainly an important step.
However, the symmetries of interest (e.g., those of this paper) are often more complex, 
involving not just space and time, but the dynamical variables as well.


\subsection*{Outline}
\label{sec:outline}

The remainder of the paper is organized as follows.
The underlying theory is reviewed in Sec.~\ref{sec:theory}, including aspects of Lie symmetries as applied to DEs, the FJet method,
and the approach for computing a constant of motion.
In Sec.~\ref{sec:1Dcase}, exact solutions are summarized for all three cases of damping for the 1D oscillator, with details being relegated to appendices.
Figures are also provided to demonstrate how a trajectory remains on a level set of a contour plot of the constant, as it should.
Following that, the matter of the integrating factor is discussed in Sec.~\ref{sec:intfactors}.
This is needed in general to compute a constant, but a way was found to first avoid needing it, and then to later compute it anyway.
Next, an interpretation of the computed constant is given in Sec.~\ref{sec:interp}.
There it is shown that the existence of the constant can be understood as being due to the conservation of energy for the {\em total} system (i.e., oscillator plus dissipative environment).
In Sec.~\ref{sec:2Dcase}, the 2D harmonic oscillator constants are found for the four cases given by whether it is damped or undamped, and isotropic or anisotropic.
Subsequent subsections consist of a verification, a computation of the constant for the commensurate cases, a generalization of angular momentum, and a generalization of the constant to arbitrary dimensions.
Following the final remarks in Sec.~\ref{sec:final}, appendices are given for details of calculations of the constants, as well as
alternative approaches for computing them.
Finally, various calculations that further explain what is presented here can be found in the Supplement.

\section{Theory}
\label{sec:theory}

As stated in the Introduction, it is natural to combine Lie symmetry techniques with FJet when pursuing constants of motion,
as they both relate to small changes in dynamical system variables.
Also, Lie symmetry techniques for differential equations (DEs) were in part developed to determine such constants.

\subsection*{FJet Approach}
\label{sec:FJet}

As introduced by the author \citep{MFZ-FJet-arxiv-current}, the FJet approach models the dynamics of a system using time series data.
In this approach, the focus is on modeling small changes in the phase space variables.
In the example of interest, the variables are the position ($u$)
and velocity ($v = du/dt$, with $t$ being the time), and differences of each would be formed over a small time step $\epsilon$ as
\begin{align*}
\Delta u & = u(t + \epsilon ) - u(t) \\
\Delta v & = v(t + \epsilon ) - v(t) \, .
\end{align*}
(Specifics related to smoothing the data are discussed in \citep{MFZ-FJet-arxiv-current}.)

The next step is to use an ML algorithm to model these small changes as $h_1$, $h_2$, which could be written as
\begin{subequations}
\begin{align*}
\Delta u & = h_1 (u,v; \epsilon) \\
\Delta v & = h_2 (u,v; \epsilon) \, ,
\end{align*}
\label{eqn:Delta_h}
\end{subequations}
where autonomous dynamics are assumed.
While any number of ML algorithms could be used for these models (e.g., NNs), it is especially
convenient to use one called {\em feature regression}; it is similar to polynomial regression,
except that it allows for more general functions.  
As an additional bonus, the functions used in feature regression are closely related to an expansion of numerical integration schemes, 
such as Runge-Kutta (RK) (cf. Ch.2 of \citep{Hairer-1987}, or App.B of \citep{Ince-1956}).
Thus, FJet (via feature regression) allows for a principled determination of the feature space (i.e., the associated functions), 
as well as a quantification of the uncertainties.  
These two benefits have been long sought after in this type of approach.

Another technique introduced by the author was to study the model parameters via regression
in the limit of $\epsilon \rightarrow 0$, and so obtain an accurate determination of the underling DE.
In summary, after starting from time-series data, it is possible to determine analytic expressions for the models $h_1$, $h_2$ (cf. \citep{MFZ-FJet-arxiv-current}).
In particular, these expressions will be used for the damped oscillator (in 1D and 2D)
\begin{subequations}
\begin{align}
 h_1 (u,v; \epsilon) & =  v \epsilon + {\cal O}(\epsilon^2) \\
 h_2 (u,v; \epsilon) & = (-\omega_0^2 u -2\gamma v ) \epsilon + {\cal O}(\epsilon^2) \, .
\end{align}
\label{eqn:HO_h1_h2}
\end{subequations}
These were found using a ML technique (i.e., FJet) in the underdamped case, but apply generally to this model.
Throughout this paper, the interest will mainly be on $h_1/\epsilon$ and $h_2/\epsilon$ as $\epsilon \rightarrow 0$.

\subsection*{Lie Symmetries}
\label{sec:Lie}

Sophus Lie \citep{Lie-1881,Olver-1993,Olver-1995,Stephani-1989,Hydon-2000,Bluman-1989} introduced the idea of using continuous symmetries in differential equations (DEs)
in order to better understand and solve them.
The symmetry may be constructed to operate on just $\{ t,u \}$ as a point symmetry, 
or may include higher order derivatives of $u$ as well (see ch.~11,12 in \citep{Stephani-1989}).
Assuming a symmetry involving $\{ t,u,v \}$, a continuous change of the variables may be written as 
$\bar{t} = f_1(t,u,v; \epsilon)$, $\bar{u} = f_2(t,u,v; \epsilon)$, and $\bar{v} = f_3(t,u,v; \epsilon)$.
where the $f_i$ ($i=1,2,3$) are analytic functions in the parameter $\epsilon$.
In particular, as $\epsilon \rightarrow 0$, the variables ($\bar{t},\bar{u},\bar{v}$) become ($t,u,v$).
In other contexts, the motivation behind using these symmetries is to reduce the order of the DE, making it more solvable.

This symmetry approach has also been used to focus on invariants ${\cal I} (\bar{t},\bar{u},\bar{v},... )$.
Under the mentioned transformations, their invariance may be expressed as
\begin{align*}
0 & = \frac{ d{\cal I} }{d\epsilon} \bigg\rvert_{\epsilon=0} 
   = \left\{ \xi\frac{\partial}{\partial \bar{t} } + \eta\frac{\partial}{\partial \bar{u} } + \eta^{(1)}\frac{\partial}{\partial \bar{v} }  + ... \right\} {\cal I} \bigg\rvert_{\epsilon=0} 
\end{align*}
with
\begin{align*}
\xi \equiv \frac{\partial \bar{t} }{ \partial \epsilon }, \, \eta \equiv \frac{\partial \bar{u} }{ \partial \epsilon }, \, \eta^{(1)} \equiv \frac{\partial \bar{v} }{ \partial \epsilon } 
\end{align*}
evaluated at $\epsilon=0$.
The method of characteristics provides for its solution via
\begin{align}
\frac{dt }{\xi} = \frac{du}{\eta} = \frac{dv }{\eta^{(1)}} = \dots \, .
\label{eqn:methodofchar}
\end{align}

Another useful technique that has been used in this context is to implement a coordinate change (e.g., \citep{Olver-1995,Hydon-2000})
to the variables $(r_1, r_2, ..., s)$, 
where the variable $s$ measures the progress along an orbit
due to the symmetry operation.  The $r_i$ emerge as constants of integration from the above equations.

The above ideas of Lie symmetries applied to DEs will play a central role in understanding
and deriving the constants of motion in conjunction with FJet.

\subsection*{Constants of Motion}
\label{sec:constants}

FJet involves computing machine-learned models $h_1$, $h_2$, which represent independent, small updates to $u,v$ over a small time step $\epsilon$;
this $\epsilon$ plays the same role as the $\epsilon$ in the previous section.
Thus, the implied variable transformation from FJet is
\begin{align*}
\bar{t} & = t + \epsilon \\
\bar{u} & = u + h_1 (u,v; \epsilon) \\
\bar{v} & = v + h_2 (u,v; \epsilon) \, .
\end{align*}
Normally, the symmetry resulting from an evolution of the model is considered trivial, since it doesn't relate different solution curves to each other,
which could help simplify the DE.
However, in the present context that is not the goal.
Instead, here the focus is on determining constants of motion using machine-learned models, and toward that end they are indeed useful.

Using the above definition of the symmetry group, the following expressions can be written for small $\epsilon$
\begin{align*}
\eta \approx h_1 / \epsilon \, , \,  \eta^{(1)} \approx h_2 / \epsilon \, .
\end{align*}
Using these in conjunction with Eq.~\ref{eqn:methodofchar} leads to
\begin{align*}
0 = -(h_2/\epsilon) \, du + (h_1/\epsilon) \, dv \, .
\end{align*}
Integrating this expression will in principle yield a constant (call it $r$),
which would seem to imply that that expression could be identified as $dr$.
However, that is only possible in general when that expression is {\em exact}; in other cases an integrating factor ($\rho$) must be used.
That is,
\begin{align*}
dr = \rho \left[ -(h_2/\epsilon) \, du + (h_1/\epsilon) \, dv \right] \, .
\end{align*}
Of course, the task still remains of determining $\rho$; it will be addressed in Sec.~\ref{sec:intfactors}.

For numerical computations, one also has to choose an initial point $(u_0,v_0)$ to begin integrating from, 
as well as the initial value $r_0$ (cf. App.~\ref{sec:app_over}).
\begin{align}
r(u,v) = r_0 + \int^{(u,v)}_{(u_0,v_0)} \rho \left[ -(h_2/\epsilon) \, du + (h_1/\epsilon) \, dv \right] \, .
\label{eqn:r_integ_init}
\end{align}
For analytic work, it's understood that $r_0$ cancels out the contribution from the lower limit, and one can simply write
\begin{align*}
r(u,v) = \int^{(u,v)} \rho \left[ -(h_2/\epsilon) \, du + (h_1/\epsilon) \, dv \right] \, .
\end{align*}
That is, the constant of integration is identified as $r(u,v)$.
In addition, in both cases it's necessary to be wary of natural boundaries formed by divergences while integrating (cf. Appendix \ref{sec:app_over}).

\section{1D Oscillator}
\label{sec:1Dcase}

As stated earlier, the starting point for this paper is a dynamical model (Eq.~\ref{eqn:HO_h1_h2}) that was obtained using FJet \citep{MFZ-FJet-arxiv-current}
with full parameter dependencies, and applies to the three types of solutions of the 1D oscillator.
In the small-$\epsilon$ limit this reproduces the usual equation of motion for a 1D damped harmonic oscillator, which is
\begin{equation*}
\ddot{u} + 2\gamma \dot{u} + \omega_0^2 u = 0 \, ,
\end{equation*}
where $\omega_0$ is the natural frequency, $\gamma$ is the damping coefficient, and $\dot{u} = du/dt = v$, as usual.
Note that $\gamma \geq 0$ and $\omega_0 > 0$.

In the following subsections the model along with Eq.~\ref{eqn:methodofchar}
will be integrated to yield constants of motion; details are relegated to the appendices.

\subsection{Underdamped Case ($\gamma < \omega_0$)}
\label{sec:1Dcase-under}

\begin{figure}
\includegraphics[scale=0.30]{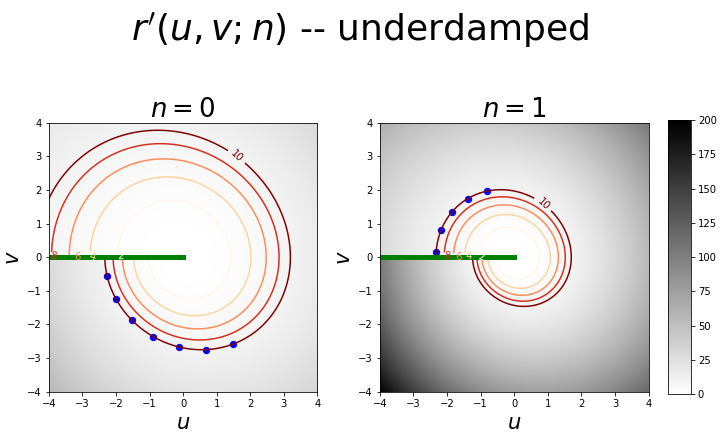}
\caption{ 
This figure displays the first two sheets ($n=0,1$) for $r'(u,v;n) = e^r$, using $r$ from Eq.~\ref{eqn:r_under_main}; each is a combination contour/heatmap plot.
The parameters used were $\omega_0 =1$ and $\gamma = 0.1$ (i.e., the underdamped case).
The horizontal green line indicates a branch cut.
The reddish-colored curves represent level sets of $r'$,
and the blue dots represent a sampling of the trajectory computed using the exact solution.
Note the blue dots stay on a single value of $r'$, as they should.
The shading and contour values are consistent between the two plots.
}
\label{fig:r_under}
\end{figure}

In the underdamped case, it is shown in App.~\ref{sec:app_under} that by diagonalizing the equation of motion (in matrix form),
and then integrating it, the following constant of motion results (see Supplement)
\begin{align}
r(u,v;n)
& = \log [ \omega^2 u^2 + ( \gamma u + v )^2 ] -  2 \frac{\gamma}{\omega} ( \phi - 2\pi n) \, ,
\label{eqn:r_under_main}
\end{align}
with
\begin{align*}
\omega & = \sqrt{ \omega_0^2 - \gamma^2 } \\
\tan \phi & = (\gamma u + v) / (\omega u) \, , \\
\end{align*}
where ``log" denotes the natural logarithm, $\phi$ lies in a $2\pi$-interval, and $n \in \mathbb{Z}$ is the Riemann sheet number.
In the Supplement it is verified with the exact solution (Eq.~\ref{eqn:exact1D_under}) that $\phi$ decreases as time increases; 
continuity is then assured by increasing $n$ when crossing the branch cut (cf. Fig.~\ref{fig:r_under}).

For visualization purposes, the constant $r' = e^r$ (with $r$ defined by Eq.~\ref{eqn:r_under_main}) is plotted in Fig.~\ref{fig:r_under}, using the sheets $n=0$ and $n=1$, with parameters $\omega_0=1$ and $\gamma = 0.1$.
In addition, values of $u,v$ from the exact solution are sampled at time steps of $0.3$,
starting from the initial condition $(u_0,v_0) = (1.5, -2.5827)$, and are shown as blue dots in the figure.
When these values are substituted into Eq.~\ref{eqn:r_under_main}, they predict the value $r' \approx 10$
Note that these dots lie on a contour of value $10$ {\em in both plots}, as they are expected to.
This test of the constancy of $r$ (or $r'$) along a trajectory is preferred by the author; another test is to simply plot $r$ versus time using trajectory data.

The branch cut (i.e., green line) limits the beginning and end of a trajectory in the graph.
Thus when a trajectory reaches a branch cut (in a clockwise fashion), it jumps to the next highest sheet.
When crossing the branch cut from below (above) the branch cut, the sheet number increases (decreases).
Note that these curves replace the elliptical energy contours used to characterize an undamped oscillator.
In the limit $\gamma \rightarrow 0$, the energy is recovered from $r'/2$.

In addition, it can be verified that $r$ is indeed a constant, using the exact solution
\begin{align}
u(t) & = e^{-\gamma t} \left[ A \cos \omega t + B \sin \omega t \right] \, ,
\label{eqn:exact1D_under}
\end{align}
with the constants $A = u_0$, $B=(v_0+\gamma u_0)/\omega$, $u(0) = u_0$ and $v(0) = v_0$.
Upon substitution into $r$, it can be shown that the time-dependence cancels out (see Supplement).

\subsection{Overdamped Case ($\gamma > \omega_0$)}
\label{sec:1Dcase-over}

\begin{figure}
\includegraphics[scale=0.30]{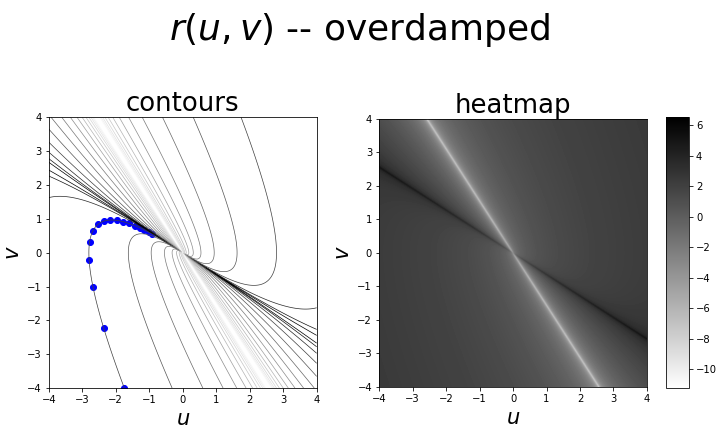}
\caption{ 
In this figure for the overdamped case (with $\omega_0 =1$, $\gamma=1.1$), a contour plot of $r$ (from Eq.~\ref{eqn:r_over_main}) is shown on the left,
and a heatmap on the right.  On the left plot, blue dots signify a sampled trajectory at time steps of $0.2$.
Note the blue dots stay on a single value of $r$, as they should.
In the right plot, the heatmap reveals the location of the positive/negative singularities along the two lines.
For graphing purposes, the minimum value allowed for $|\zeta u \pm (\gamma u + v) |$ was $0.001$.
}
\label{fig:r_over}
\end{figure}

In the overdamped case, it is similarly shown in App.~\ref{sec:app_over} that by diagonalizing the equation of motion and then integrating it, a constant of motion can be obtained
\begin{align}
r(u,v) = & -( \zeta + \gamma ) \log | \zeta u + (\gamma u   + v) | \nonumber \\
    & - (\zeta - \gamma ) \log |  \zeta u - ( \gamma u + v) | \, ,
\label{eqn:r_over_main}
\end{align}
where $\zeta = \sqrt{ \gamma^2 - \omega_0^2 }$.
In Fig.~\ref{fig:r_over} this constant is plotted using $\omega_0=1$ and $\gamma = 1.1$.
These two plots were given separately in this case since the contour lines tend to accumulate where the heatmap has singularities.
Of note is that there is no longer a rotation about the origin, as there was for the underdamped case.
As before, values of $u,v$ from the exact solution are sampled at time steps of $0.2$,
starting from the initial condition $(u_0,v_0) = (-1.75, -3.99)$, and are shown as blue dots.
When these values are substituted into Eq.~\ref{eqn:r_over_main}, they predict the value $r \approx 2$
Note that these dots lie near a contour of value $2$, as they should.

Similar to the underdamped case, one can substitute in the following exact solution for $u(t)$ 
\begin{align*}
u(t) & = A e^{-\gamma t + \zeta t} + B e^{-\gamma t -\zeta t} \\
A & = [ \zeta u_0 + ( \gamma u_0 + v_0) ] / (2 \zeta) \\
B & = [ \zeta u_0 - ( \gamma u_0 + v_0) ]  / (2 \zeta)
\end{align*}
into Eq.~\ref{eqn:r_over_main} to show it is indeed independent of time.
As one would expect, the constant takes the same form, with $u,v$ replaced by $u_0,v_0$.

\subsection{Critically Damped Case ($\gamma = \omega_0$)}
\label{sec:1Dcase-critical}

\begin{figure}
\includegraphics[scale=0.30]{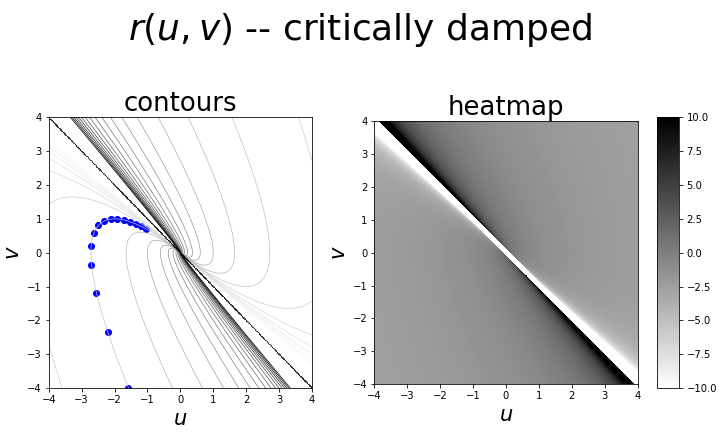}
\caption{ 
In this figure for the critically damped case (with $\omega_0 = \gamma=1$), a contour plot of $r$ (from Eq.~\ref{eqn:r_critical_main}) is shown on the left,
and a heatmap on the right.  On the left plot, blue dots signify a sampled trajectory at time steps of 0.2.
Note the blue dots stay on a single value of $r$, as they should.
In the right plot, the heatmap reveals the location of the positive/negative singularities where $| \gamma u + v|$ equaled zero; its minimum value was limited to $0.01$ for graphing purposes.
}
\label{fig:r_crit}
\end{figure}

In the critically damped case, it is shown in App.~\ref{sec:app_crit} the constant of motion is
\begin{align}
r(u,v) = & \log | \gamma u   + v | + \left( \frac{ \gamma u }{ \gamma u + v } \right) \, .
\label{eqn:r_critical_main}
\end{align}
This constant is plotted in Fig.~\ref{fig:r_crit} using $\omega_0= \gamma = 1$.
In the left plot is a contour plot along with blue dots representing a trajectory sampled at time steps of 0.2, beginning from $(u,v)=(-1.58,-3.99)$.
These dots can be seen to lie along a level set of the contour plot (as they should), at value $r \approx -2$.
In the right plot is shown a heatmap of $r$, displaying singular behavior along the line $\gamma u + v = 0$ (as expected).

Finally, it is easily verified that upon substituting the exact solution
\begin{align*}
u(t) & = (A + Bt ) e^{-\gamma t }
\end{align*}
into Eq.~\ref{eqn:r_critical_main}, with $A = u_0$, $B = \gamma u_0 + v_0$, $u_0 = u(0)$, and $v_0 = v(0)$,
there is no remaining time-dependence.

\section{Integrating Factors}
\label{sec:intfactors}

As discussed earlier, an integrating factor ($\rho$) is in general needed to obtain the constant $r$.
However, in appendices \ref{sec:app_under} and \ref{sec:app_over} it was nevertheless computed without one.
This was possible because the variables $(u,v)$ were transformed to $(\tilde{u}, \tilde{v})$ so that the equation of motion had a diagonal matrix representation.
Using the new variables, the equations to be integrated generically appeared as 
\begin{align}
0 = f( \tilde{u} ) \, d\tilde{u} - g( \tilde{v} ) \, d\tilde{v} 
\label{eqn:intfac0}
\end{align}
for some functions $f$ and $g$.
For this to be interpreted as arising from a differential of a constant $\Phi(\tilde{u},\tilde{v})$,
one could interpret partial derivatives as $\Phi_{\tilde{u}} = f(\tilde{u})$ and $\Phi_{\tilde{v}} = -g(\tilde{v})$.
Also, since the mixed partials are equal (i.e., $\Phi_{ \tilde{u}\tilde{v} } = 0 = \Phi_{ \tilde{v}\tilde{u} }$),
it follows that Eq.~\ref{eqn:intfac0} is exact, and an integrating factor isn't needed.
Hence, a constant $r(u,v) = \Phi(\tilde{u},\tilde{v})$ can then be written as
\begin{align*}
r(u,v) = \int [ f( \tilde{u} ) d\tilde{u} - g( \tilde{v} ) d\tilde{v} ] \, ,
\end{align*}
as was done in those appendices.

Even though it was possible to avoid the need for an integrating factor (in Apps.~\ref{sec:app_under} and \ref{sec:app_over}),
it is still of interest to know what it is.
Since the constant $r$ is already known, it can be computed by forming its differential,
and then identifying the common factor as the integrating factor $\rho$, that is,
\begin{align*}
dr & = r_u du + r_v dv \\
& = \rho \, [ \, (\omega_0^2 u + 2\gamma v) du + v dv \, ]  \, .
\end{align*}
Indeed, this was easily done for all three damping cases, and result could be summarized into a single expression
\begin{align*}
\rho
& = \frac{ 1 }{ (\omega_0^2 - \gamma^2) u^2 + (\gamma u + v)^2 }  \, .
\end{align*}
Overall constant factors are unimportant and were ignored.
Finally, the reader is reminded that an integrating factor is not unique (e.g., see p.27 in \citep{Ince-1956}).

\section{Interpretation of $r$}
\label{sec:interp}

Given that $r$ is a constant, it is natural to ask what it represents.
In the undamped case ($\gamma=0$) it is related to the energy, but it is not yet clear what it is in general.

Beginning from Eq.~\ref{eqn:r_integ_init}, perform an integration along a trajectory $\Gamma$ formed by time-evolution, 
starting from initial conditions $(u_0,v_0)$ where $r \equiv r_0$.
Of interest is the change in $r$ along this path (i.e. $\delta r= r(u,v) - r_0$), and is computed as
\begin{align*}
\delta r 
& = \int_\Gamma \rho \left[ -(h_2/\epsilon) \, du + (h_1/\epsilon) \, dv \right] \\
& = \int_\Gamma \rho \left[  (\omega_0^2 u + 2 \gamma v) \, du + v \, dv \right] \\
& = \int_\Gamma \rho \left[  dE + dW \right] \, ,
\end{align*}
where $E = \frac{1}{2} (\omega_0^2 u^2 + v^2)$ is the energy, and $W = \int^u 2 \gamma v \, du'$ is the work done on the oscillator by the dissipative environment.
(Note that $dW = 2 \gamma v^2 dt > 0$.)
Thus, $dE + dW$ is the change in the {\em total} energy for the oscillator + environment over the time step $\epsilon$.
However, it can be seen (by several means) that the sum of these two changes is zero for this class of systems.
The most straightforward way is to substitute in $dv = -(\omega_0^2 u + 2\gamma v) dt$ from Eq.~\ref{eqn:HO_h1_h2}, use $vdt=du$, and so obtain $dE + dW=0$.
Another way is to use the definitions of $h_1$ and $h_2$ in terms of updates (cf. Eq.~\ref{eqn:Delta_h}), to write it as
\begin{align*}
\delta r 
& \approx \int_\Gamma \rho \left[ -(\Delta v/\epsilon) \, du + (\Delta u/\epsilon) \, dv \right] \, .
\end{align*}
For small time steps ($\epsilon$), $\Delta u$ and $\Delta v$ approximate $du$ and $dv$, and the integrand approaches $0$.
Finally, one could also simply note that the variable change of $(u,v) \rightarrow (r,s)$ was {\em constructed} so that $r$ would be a constant (see discussion in Sec.~\ref{sec:Lie}).

Thus, for this example of a 1D damped oscillator, it's been shown that the existence of the constant $r$ is a reflection of the fact that the 
energy of the oscillator plus the accumulated work done by its dissipative environment is constant.
Also, it should be clear that this example belongs to a wider class of models that exhibit this property.
For example, one might consider the equation of motion for a general gradient-based force and general velocity-dependent damping.
In that case \citep{MFZ-unpub-2023} it's also true that $\delta r= 0$.
Of course, the property of conservation of energy is universal, so finding it manifested as a constant in this way is really just a matter of properly defining the equations of motion.

\section{2D Oscillator}
\label{sec:2Dcase}

The 2D harmonic oscillator offers an interesting testing ground, 
as it carries additional symmetry compared to the 1D oscillator,
and it bears some similarity to planetary motion (for example).
For this 2D case the focus will be on the \underline{underdamped} case for each direction.
There are actually six unique combinations of under/over/critically damped cases for the two directions,
but they are similar to that already presented and so are left as an exercise for the reader.

The equations of motion are a generalization of the 1D case
\begin{equation*}
\ddot{u}_k + 2\gamma \dot{u}_k + \omega_{k0}^2 u_k = 0 \, ,
\end{equation*}
where $k=1,2$ (which will be used through this section).
Since the $u_k$ are uncoupled, it may be assumed data was already collected for both $u_1$ and $u_2$ in the same manner as before,
and they were both then analyzed with FJet, yielding a pair of dynamical models in an obvious generalization of the 1D case.
Hence, the analysis may now continue with these definitions
\begin{align*}
\omega_k & = \sqrt{\omega_{k0}^2 - \gamma^2} \\
\lambda_{k\pm}  & = - \gamma \pm i\omega_k  \\
\tilde{u}_k & = \frac{1}{2\omega_k} [ \omega_k u_k - i( \gamma u_k + v_k) ] \\
\tilde{E}_k & = \frac{1}{2} [ (\omega_k u_k)^2 + (\gamma u_k + v_k)^2 ]  \, ,
\end{align*}
and the phases
\begin{subequations}
\begin{align}
\tan \phi_k & = ( \gamma u_k + v_k) / (\omega_k u_k) \label{subeqn:tanphi-k} \, ,
\end{align}
\label{subeqn:2D-all-phi}
\end{subequations}
which are interpreted in a similar manner as in Sec.~\ref{sec:1Dcase-under}.
In addition, $\tilde{v}_k = \tilde{u}_k^*$ (i.e., the complex conjugate),
and $\tilde{E}_k$ is a pseudo-energy which in the limit $\gamma=0$ equals the actual energy $E_k$ (for the $k$-th component in each case)
\begin{align*}
E_k = \tilde{E}_k \big\rvert_{\gamma = 0} \, .
\end{align*}
Again using Eq.~\ref{eqn:methodofchar} (as was done in App.~\ref{sec:app_under}), it follows
\begin{equation}
\frac{ d \tilde{u}_1 }{\lambda_{1+} \tilde{u}_1} = \frac{ d \tilde{v}_1 }{\lambda_{1-} \tilde{v}_1} = dt
= \frac{ d \tilde{u}_2 }{\lambda_{2+} \tilde{u}_2} = \frac{ d \tilde{v}_2 }{\lambda_{2-} \tilde{v}_2} \, .
\label{eqn:2D-5parts}
\end{equation}
There are now ${4\choose 2} = 6$ ways of choosing a pair of terms for integration; the $dt$ term is ignored.
If either the first two terms or last two terms are chosen, a constant very similar to Eq.~\ref{eqn:r_under_main} is obtained: i.e., $r(u_k, v_k; n)$ for sheet n.
Thus, in the subsections to follow, the cases of interest will involve an integration between, for example, $\tilde{u}_1$ and $\tilde{v}_2$.
Also, various special cases are considered, such as undamped cases ($\gamma=0$),
isotropic cases ($\omega_{10}=\omega_{20}$), as well as general anisotropic cases ($\omega_{10} \neq \omega_{20}$), both commensurate and incommensurate.

\subsection{Undamped, Anisotropic}

In this undamped ($\gamma=0$), anisotropic case, {\em no} special relationship between $\omega_{10}$ and $\omega_{20}$ is assumed.
When $\omega_{10} \neq \omega_{20}$, this is a non-central force problem and angular momentum is not conserved.
In this general case, the parameters become $\omega_k = \omega_{k0}$, $\lambda_{k\pm} = \pm i\omega_{k0}$.

At this point, there are several options to choose from in deciding which pair of terms of Eq.~\ref{eqn:2D-5parts} should be integrated.
If the $\tilde{u}_1,\tilde{v}_1$,  terms are chosen, the result will be the same form as Eq.~\ref{eqn:r_under_main} for the 1D case, except that now $\gamma=0$.
Thus, it simply derives that $E_1$ is a constant.   Likewise, if the $\tilde{u}_2,\tilde{v}_2$ terms are chosen, the result shows that $E_2$ is constant.
A more interesting result follows from integrating the $\tilde{u}_1$ and $\tilde{v}_2$ terms of Eq.~\ref{eqn:2D-5parts}, leading to the constant
\begin{align}
r & = - i\omega_{20} \log  [ \omega_{10} u_1 - iv_1 ]  - i\omega_{10} \log [ \omega_{20} u_2 + i v_2] \, ,
\label{eqn:r_aniso_2}
\end{align}
where $u_k,v_k$-independent terms have been dropped.
The real and imaginary parts of $r$ are defined as $C_R$ and $C_I$, that is
\begin{align*}
C_R & =  - i\omega_{20} \log \left(  e^{ -i\phi_1} \right) - i\omega_{10} \log \left( e^{i\phi_2} \right) \\
C_I  & = - \omega_{20} \log \sqrt{ 2E_1 } - \omega_{10} \log \sqrt{ 2E_2 }  \, ,
\end{align*}
where in this case $\tan \phi_k = v_k / (\omega_{k0} u_k )$ (cf. Eq.~\ref{subeqn:tanphi-k}).
Note that $C_I$ is already constant since the $E_k$ are constant.
In a similar manner to the treatment from before, Riemann sheets are needed, and the constant is
\begin{align}
C_R & = \omega_{10} \phi_2 - \omega_{20} \phi_1  - 2\pi ( n_2 \omega_{10} - n_1 \omega_{20}) \, ,
\label{eqn:CR-aniso-undamped}
\end{align}
where $n_k \in \mathbb{Z}$ and the $\phi_k$ are confined to a $2\pi$-interval.

The content of Eq.~\ref{eqn:CR-aniso-undamped} has recently been derived by other means in \citep{ASinha-2023b}, 
except that the $\phi_k$ were generically expressed via ``$\tan^{-1}$",
without mention of the Riemann sheets 
or the ramifications covered here in Sec.~\ref{sec:2D-interp} to \ref{sec:2D-wedge}.
A more complete comparison is given in Sec.~\ref{sec:comparison}.

\subsection{Undamped, Isotropic}
\label{sec:2D-undamped-iso}

In this isotropic case, one can simply take the results from the previous (anisotropic) case and set 
$\omega_{10} = \omega_{20} \equiv \omega_{0}$.
The constant $C_R$ will now be referred to as $C_R^{iso}$; likewise, $E_1$, $E_2$ will also change.
Again, the only constant of interest is the real part, and in this isotropic case, it becomes
\begin{align}
C_R^{iso} & =  - i\omega_0 \log \left(  e^{ -i\phi_1} \right) - i\omega_0 \log \left( e^{i\phi_2} \right) \\
& = \omega_0 \left[ \phi_2 - \phi_1  - 2\pi ( n_2 - n_1) \right]  \, ,
\label{eqn:CR-iso}
\end{align}
where $n_k \in \mathbb{Z}$ and $\tan \phi_k = v_k / (\omega_0 u_k )$ (cf. Eq.~\ref{subeqn:tanphi-k}).
Motivated from results using the exact solution (see Supplement), the following expressions are computed 
\begin{subequations}
\begin{align}
\sin (C_R^{iso} / \omega_0 ) 
& = \sin \phi_2 \cos \phi_1 - \cos \phi_2 \sin \phi_1  \nonumber \\
& = \frac{ \omega_0 }{ 2\sqrt{E_1 E_2} } \left( u_1 v_2 - u_2 v_1 \right) \label{subeqn:sin-undamped-iso} \\
\cos (C_R^{iso} / \omega_0 ) 
& = \cos \phi_2 \cos \phi_1 + \sin \phi_2 \sin \phi_1  \nonumber \\
& = \frac{1}{2 \sqrt{ E_1 E_2} } \left( \omega_0^2 u_1 u_2 + v_1 v_2 \right) \label{subeqn:cos-undamped-iso}  \, .
\end{align}
\label{subeqn:2D-undamped-iso}
\end{subequations}
Since $C_R^{iso}$, $E_1$, and $E_2$ are each constants, it follows that $(u_1 v_2 - v_1 u_2)$ and $(\omega_0^2 u_1 u_2 + v_1 v_2)$
are constants as well.  These are of course proportional to the constants $S_3$ and $S_1$ on p.~424 in \cite{Goldstein-book-2nd},
with the former being familiarly recognizable as the angular momentum.
In addition, in this format it is easy to see that they're related as
\begin{align*}
4 E_1 E_2 & = \omega_0^2 \left( u_1 v_2 - v_1 u_2 \right)^2 + \left( \omega_0^2 u_1 u_2 + v_1 v_2 \right)^2 \, .
\end{align*}
Finally, it is noted that the simplicity of this isotropic case also permits a simpler rewriting.
Equation \ref{eqn:r_aniso_2} may instead be written as
\begin{align*}
r & = - i\omega_0 \log [ C_R' + i C_I' ] + \dots \\
C_R' & = \omega_0^2 u_1 u_2 + v_1 v_2 \\
C_I' & = \omega_0 ( u_1 v_2 - u_2 v_1 ) \, ,
\end{align*}
where again $\dots$ indicate $u_k,v_k$-independent terms.

\subsection{Damped, Anisotropic/Isotropic}

Following the pattern for the undamped cases, the constants for the anisotropic case are computed first,
and then the isotropic case is derived from it by setting the frequencies equal.

\subsubsection*{Anisotropic}
\label{sec:2D-damped-aniso}

Constants of motion were found by integrating the $\tilde{u}_1,\tilde{u}_2$ pairs and then the $\tilde{u}_1,\tilde{v}_2$ pairs from Eq.~\ref{eqn:2D-5parts}.
Adding and subtracting those results led to the constants
\begin{subequations}
\begin{align}
C_A & =  \gamma \log \left( \frac{ \tilde{E}_2 }{ \tilde{E}_1} \right) \nonumber \\
C_B & = \omega_1 \phi_2 - \omega_2 \phi_1 - 2 \pi (  n_2 \omega_1 - n_1 \omega_2 )  \label{subeqn:CB}  \, ,
\end{align}
\label{subeqn:2D-damped-aniso}
\end{subequations}
where $\tan \phi_k = ( \gamma u_k + v_k) / (\omega_k u_k)$, and $\phi_k$ is confined to a $2\pi$-interval.
Interestingly, while $\tilde{E}_k$ is {\em not} a constant in this damped case,
the ratio $\tilde{E}_2 / \tilde{E}_1$ is.  (The exact solution shows that $\tilde{E}_k \propto e^{-2\gamma t}$,
with no other time dependence.)
The constant $C_B$ is also very interesting in that it is of the same form as in the undamped case (Eq.~\ref{eqn:CR-aniso-undamped}),
which was unexpected.

\subsubsection*{Isotropic}

The results found above for the damped, anisotropic case can be applied here to the damped isotropic case,
by setting $\omega_1 = \omega_2 \equiv \omega$.
The constant $C_A$ is still valid here, except that the pseudo-energies $\tilde{E}_k$ are slightly modified according to the $\omega_{k0}$.
Also, the constant $C_B$ can be written, in a manner similar to what was done for the undamped, 
isotropic case (Eq.~\ref{subeqn:2D-undamped-iso}), that is
\begin{subequations}
\begin{align}
\sin (C_B / \omega ) 
& = \frac{ \omega }{ 2 \sqrt{ \tilde{E}_1 \tilde{E}_2} }  \left( u_1 v_2 - u_2 v_1 \right) \label{subeqn:sin-damped-iso} \\
\cos (C_B / \omega ) 
& = \frac{1}{2 \sqrt{ \tilde{E}_1 \tilde{E}_2} } \left( \omega^2 u_1 u_2 + v_1 v_2 \right) \label{subeqn:cos-damped-iso} \, ,
\end{align}
\label{subeqn:2D-damped-iso}
\end{subequations}
where now $\tan \phi_k = ( \gamma u_k + v_k) / (\omega u_k)$.
Note that using the identity alluded to previously, one can write
\begin{align*}
4 \tilde{E}_1 \tilde{E}_2 & = \omega^2 \left( u_1 v_2 - v_1 u_2 \right)^2 + \left( \omega^2 u_1 u_2 + v_1 v_2 \right)^2 \, .
\end{align*}
However, what is interesting is that both the left and right-hand sides decay as $e^{-2\gamma t}$.

\subsection{Verification}

There are several ways to verify that the previously computed constants are indeed time-independent.
For the undamped case, the Hamiltonian is
\begin{align*}
H_N = \sum_{k=1}^N \frac{1}{2} ( \omega_{k0}^2 u_k^2 + v_k^2 ) \, .
\end{align*}
Using $C_R$ from Eq.~\ref{eqn:CR-aniso-undamped}, it is easily shown in the Supplement that
\begin{align*}
\{ C_R, H_2 \} = 0 \, ,
\end{align*}
proving it is a conserved quantity.  
It is also shown there that using the exact solution
\begin{align*}
u_k = A_k \cos ( \omega_{k0} t - \beta_k)
\end{align*}
in the isotropic, undamped limit (with $\omega_{10} = \omega_{20} = \omega_0$), that
\begin{align*}
& \omega_0^2 u_1 u_2 + v_1 v_2  \longrightarrow \omega_0^2 A_1 A_2 \cos ( \beta_1 - \beta_2 ) \\
& \omega_0 ( u_1 v_2 - u_2 v_1 ) \longrightarrow \omega_0^2 A_1 A_2 \sin ( \beta_1 - \beta_2 ) \\
& \omega_{10} \phi_2 - \omega_{20} \phi_1 \longrightarrow \omega_0 ( \beta_1 - \beta_2 ) \, .
\end{align*}
These exact results were the motivation for the author to compute the sin/cos in Eq.~\ref{subeqn:2D-undamped-iso}
(and also Eq.~\ref{subeqn:2D-damped-iso}).
In addition, it's also easy to show that
\begin{align*}
\tan \phi_k & = \tan ( -\omega_{k0} t + \beta_k )  
\end{align*}
leads to
\begin{subequations}
\begin{align}
\phi_k & = [ -\omega_{k0} t + \beta_k ] + 2m\pi \nonumber  \\  
\phi_k & = -\omega_{k0} t + \beta_k \label{eqn:u_exact_aniso} \, ,
\end{align}
\end{subequations}
where $[ \, \,  ]$ signifies restriction to a $2\pi$-interval, and $m$ is a Riemann sheet number.
Using the exact solution for the anisotropic, damped case, a similar expression follows (i.e., $\phi_k = -\omega_k t + \beta_k$).

\subsection{Commensurate Frequencies}
\label{sec:2D-comm}

The undamped, anisotropic 2D oscillator is known to have invariants when the frequencies are commensurate, 
which can be computed using an action angle approach (cf. p.14 in \cite{Atlee-book-v2}). 
However, the results in this paper can also be used to derive them.
In this commensurate case, one can write $\omega_{10} / \omega_{20} = a/b$, where $a,b$ are relatively prime, positive integers.
After setting $\omega_{10} = \bar{\omega} a$, $\omega_{20} = \bar{\omega} b$, with $\bar{\omega} > 0$, 
the constant (Eq.~\ref{eqn:CR-aniso-undamped}) appears as
\begin{align}
C_R / \bar{\omega}
& =  a \phi_2 - b \phi_1 - 2\pi ( a n_2 - b n_1 ) \, .
\label{eqn:C-comm}
\end{align}
The term $( a n_2 - b n_1 )$ is always an integer, and so
\begin{align*}
\sin( C_R / \bar{\omega} )
& =  \sin(a \phi_2) \cos( b \phi_1) - \cos(a \phi_2) \sin( b \phi_1) \, .
\end{align*}
Using standard trigonometric formulas for the sin/cos functions,
the expression can be simplified to a polynomial.  For example, in the case $a=1$, $b=2$, it becomes
\begin{align}
\sin( C_R / \bar{\omega} )
& = \frac{1}{ \sqrt{ 8E_1^2 E_2}  } \left[ \bar{\omega}^2 u_1^2 v_2 - v_1^2 v_2 - 4 \bar{\omega}^2 u_1 u_2 v_1 \right]  \, .
\label{eqn:commens}
\end{align}

Such constants have long been known for commensurate frequencies, but they hadn't previously been shown to be related to a constant
that had no such commensurability constraints (i.e., Eq.~\ref{eqn:CR-aniso-undamped}).
Also, a similar identity could also be found by taking the cosine of Eq.~\ref{eqn:C-comm}.
This particular example was verified in the Supplement using the exact $u_k$, where it was shown that
$\sin( C_R / \bar{\omega} ) = \sin ( \beta_2 - 2\beta_1 )$, as one would expect from Eqs.~\ref{eqn:u_exact_aniso},\ref{eqn:C-comm} evaluated at $t=0$.

Finally, an additional result is that this kind of identity also holds when there is damping.
One would begin from Eq.~\ref{subeqn:CB} and similarly set $\omega_1 = \bar{\omega} a$, $\omega_2 = \bar{\omega} b$, 
to obtain the same result as in Eq.~\ref{eqn:commens}, except now with $E_k \rightarrow \tilde{E}_k$,
and $u_k,v_k$ referring to the damped solution.

\subsection{Interpretation}
\label{sec:2D-interp}

In this section, some of the previous constants for the undamped cases are collected for a comparison.
Beginning from $C_R$ in Eq.~\ref{eqn:CR-aniso-undamped}, set
\begin{align}
\Phi & = ( \omega_{10} \phi_2 - \omega_{20} \phi_1 ) / \omega_{10} \, .
\label{eqn:Phi}
\end{align}
Recall that by taking projections with this angle, vis-\`a-vis Eq.~\ref{subeqn:2D-undamped-iso},
constants of motion could be recovered.
In general, for any ratio of $\omega_{10}/\omega_{20}$, 
it is true that the projections are $A = 2\sqrt{E_1 E_2}\cos \Phi$ and $B = 2\sqrt{E_1 E_2}\sin \Phi$.
In Table~\ref{tab:interp}, values for $B$ are also shown for the cases $\omega_{10}/\omega_{20} =1$, $1/2$,
and an arbitrary irrational number (where the general expression for $B$ is shown).
\begin{table}[b]
\caption{ \label{tab:stability-logrsq}
In all three cases $B = 2\sqrt{E_1 E_2} \sin \Phi$, but in the first two cases it's known how to simplify it down to a polynomial, as given in the table.  This suggests that for the 2D oscillator, angular momentum ($L$) is just another such polynomial, whereas $\Phi$ is a robust constant.
}
\begin{ruledtabular}
\begin{tabular}{ccr}
  $\omega_{10} / \omega_{20}$  & $B$    &  $dL/dt$ \\
\colrule
1                                              &  $\omega_{10} \left(u_1 v_2 - u_2 v_1 \right)$  &  $0$ \\
$1/2$             &  $\left(  \omega_{10}^2 u_1^2 v_2 - v_1^2 v_2 - 4\omega_{10}^2 u_1 u_2 v_1 \right) / \sqrt{2E_1} $  & $\neq 0$ \\
irrational   &  $2\sqrt{E_1 E_2}\sin \Phi$  &  $\neq 0$ \\
\end{tabular}
\end{ruledtabular}
\label{tab:interp}
\end{table}
The third column shows that angular momentum ($L = u_1 v_2 - u_2 v_1$) is only conserved when $\omega_{10} / \omega_{20} =1$; 
it is not a constant otherwise.
In comparison, $\Phi$ is a constant for all values of $\omega_{10} / \omega_{20}$.
Thus, it appears that $L$ is just another polynomial constant, not unlike that for the case $\omega_{10} / \omega_{20} =1/2$.
Another way to describe this situation is to say that the constant
\begin{align*}
C' = \frac{2}{\omega_{10} } \sqrt{ E_1 E_2 } \sin \Phi 
\end{align*}
generalizes angular momentum.
This is because it is a constant in general, and as $\omega_{10}/\omega_{20} \rightarrow 1$ it becomes equal to the angular momentum.
Finally, similar remarks can be made regarding the horizontal projection $A$.

Finally, note that $C'$ is not unique in this regard.  For example, one could use $\omega_{20}$ instead of $\omega_{10}$ as a divisor in Eq.~\ref{eqn:Phi}.
Note that when $\Phi$ is written in either manner, it limits which cases of $\omega_{10} / \omega_{20}$ can be compared, as was done in Table~\ref{tab:interp}.
(As written, one would need $\omega_{10} / \omega_{20} = 1/n$, with $n \in \mathbb{Z}^+$.)
To be able to obtain a polynomial invariant for {\em any} commensurate case, one should instead divide by $\bar{\omega}$, as was done in the previous section.
However, defining $\bar{\omega}$ formulaically is non-obvious.

\subsection{Arbitrary Dimensions}
\label{sec:2D-wedge}

The steps taken for the $2$D damped, anisotropic oscillator can likewise be taken in $N$-dimensions.
That is, the computations that were done in Sec.~\ref{sec:2D-damped-aniso} with respect to the directions $1,2$ can now be done for the arbitrary
directions $i,j$, where $1 \leq i,j \leq N$ and $i \neq j$.
Analogous computations on the $\tilde{u}_i,\tilde{u}_j$ pairs and then the $\tilde{u}_i,\tilde{v}_j$ pairs likewise lead to the constant
\begin{align}
C (i,j) 
& = \omega_i \varphi_j - \omega_j \varphi_i \, .
\label{eqn:C-general}
\end{align}
where $\varphi_k = \phi_k - 2\pi n_k$ (with $k=i,j$), and $\phi_k$ is again confined to a $2\pi$-interval. 
All of the different $i,j$ values can be encapsulated after defining the row vectors 
\begin{align*}
{\bm \omega} & = (\omega_1, \omega_2, ..., \omega_N ) \\
{\bm \varphi} & = ( \varphi_1, \varphi_2, ..., \varphi_N ) ,
\end{align*}
and Eq.~\ref{eqn:C-general} can be rewritten as an exterior product, i.e., $C = {\bm \omega} \wedge {\bm \varphi}$.
Note that it is comprised of $N(N-1)/2$ constants.
Also, using the on-shell relation of Eq.~\ref{eqn:u_exact_aniso}, it can be rewritten as
\begin{align*}
C = \bm{\varphi} \wedge \dot{ {\bm \varphi} } \, .
\end{align*}
The reader might note the similarity to angular momentum with unit mass (i.e., ${\bm r} \wedge {\dot {\bm r} }$, where ${\bm r}$ is the coordinate).
Also, it is easily shown that $C$ is constant in time (see the Supplement).
Finally, this constant has only been shown for the damped / undamped $N$-dimensional oscillator, and so the question remains
as to what degree such an elegant expression persists for more realistic systems.

\section{Comparisons}
\label{sec:comparison}

The author has presented an approach which is capable of producing {\em all} constants of motion for a dynamical
systems from a single, generic data set.  In particular, such a data set could be comprised of a single trajectory of data \citep{MFZ-unpub-2021}.
When there is no dissipation, the method recovers the traditional constants of motion (e.g., energy, momentum);
when dissipation is present, it recovers new constants of motion, as described herein.  
In addition, as demonstrated on the conserved case of the 2D anisotropic harmonic oscillator, 
a constant of motion was identified that generalizes angular momentum.

Papers \citep{SWetzel-2020,SHa-2021} displayed clever ways to exploit the embedding layer of a NN to compute constants of motion.
However, the work they presented was subject to several constraints:
(1) they could only find constants they already knew;
(2) they required specially crafted data sets;
(3) they could only find one constant at a time from a given data set.
For example, to discover angular momentum, the authors of \citep{SWetzel-2020} needed to create data in two classes, one with the same value of angular momentum, and one without.  Similarly, the authors of \citep{SHa-2021} needed to create data with stratified values of angular momentum.
Note that it is often the case that one doesn't have the flexibility to specially construct the data: for example, the data might come from an observation of a single planetary orbit.  

The author also has concerns about the use of the noise-variance (NV) loss function in \citep{SHa-2021}.
Although they argue for their approach using generating functions and other means,
it is not obvious to the author that their loss function does not influence the polynomial regression (i.e., the derived invariant).
Toward understanding this better, it would be interesting to test their invariants to see what kind of dependence they had on the parameters of their NV-loss function (i.e., $Q$, $R$).

Both \citep{YMototake-2021} and the present paper aim to use invariance under a continuous symmetry (i.e., time evolution) to determine invariants of the system.
However, they do it for conserved systems, whereas here it is done for both conserved and dissipative ones.
Also, they seek to keep the Hamiltonian invariant, whereas here the approach is to work with a general invariant.
After these similar starting points, the computational paths taken by each paper are very different, with one exception:
they use polynomial regression to model the infinitesimal changes, and here feature regression was used (cf. \citep{MFZ-FJet-arxiv-current}).
In the end, they were able to determine an invariant for some model systems, but did write that the complexity of their technique 
would limit future applications, especially if attempted on multiple invariants.
In contrast, the reader should have found that the techniques presented herein are 
relatively straightforward, and can readily handle multiple invariants.

In one of their examples, the authors of \citep{ZLiu-2021} used their method to detect symmetries in a decaying planetary orbit,
and found it could only be used for a short time.  However, as they introduced their method, it should have been general enough
to detect the new types of constants discussed in this paper (when dissipation is present).
It's not clear to the author whether this was due to their method (based on \citep{SSaremi-2019}), or their application of it.
It might also be related to assumptions of how the trajectories are embedded in the phase space, which is qualitatively different for conserved and damped oscillators.
Also, their failure to discover the true invariant might also be due to their use of symbolic regression dictionaries \citep{Udrescu-2020-AIFeynman2}, 
which in the author's opinion lacks a mathematical basis.

After the author completed work on the 1D oscillator (i.e., Secs.~\ref{sec:1Dcase}, \ref{sec:intfactors}, \ref{sec:interp}, and App.~\ref{sec:alt_solutions}),
he realized that one of his alternative solutions could be used to connect his results to previous work (although that work made no mention of damped oscillators).  
Using the variable definition of $w = \gamma u + v$ in App.~\ref{sec:alt_solutions_2}, 
one can switch to polar coordinates an obtain a logarithmic spiral solution.
As pointed out by Olver (see pages 105,136 in \citep{Olver-1993}), that set of solutions exhibit an SO(2) symmetry,
as they can be rotated into each other.  Also, with this coordinate change, the constant found here for the 1D case (Eq.~\ref{eqn:r_under_main}) 
becomes similar in appearance to that on p.136 of \citep{Olver-1993}.

After the author computed the constant $C_R$ in Eq.~\ref{eqn:CR-aniso-undamped} for the undamped anisotropic harmonic oscillator, he became aware of a recent preprint in which the same constant had already been found (cf. Eq. 27 in \citep{ASinha-2023b}).  
However, they generically casted the constant in terms of $\tan^{-1}$, skipping mention of the Riemann sheets.
Also, they didn't consider the ramifications investigated here: relation to commensurate frequencies, generalization of angular momentum, generalization to higher dimensions, and the presence of damping.
Nevertheless, their analytical approach is apparently well-suited for conserved systems with known Hamiltonians, as they demonstrated on several problems.
However, the approach presented here is also able to easily treat dissipative, non-Hamiltonian systems.
Interestingly, they also derived a constant for a Bateman pair oscillator, which contains similar elements to the constant previously found by the author for the 1D damped oscillator \citep{MFZ-MLDE-arxiv-v2}.

\section{Final Remarks}
\label{sec:final}

This paper began as part of an effort to apply ML techniques to discovering physical laws, given only time series data and perhaps some domain information.  
The idea had been to combine two techniques which studied dynamics over small variable changes (i.e., FJet and Lie symmetry techniques for DEs),
and use them to determine constants of motion.
However, during this study it was found that there existed constants of motion for {\em dissipative} systems, in addition to conservative ones.
It was surprising that such a well-studied example (1D damped harmonic oscillator) still had some secretes to reveal, and so additional attention was paid to it, as evidenced by some sections of this paper.  
It was also surprising that it would reveal the relationship of a relatively unknown constant (see Sec.~\ref{sec:2D-interp}) to angular momentum.

Finally, it may appear that this has been a purely analytical paper, without any reliance on numerical computation or ML.
However, the reader should note that the basis of this paper was an initial computation of the models $h_1$, $h_2$, which can be found using any regression-type ML model (e.g., NN).


\appendix

\section{Derivation of $r$ (underdamped)}
\label{sec:app_under}

It follows from the FJet model \citep{MFZ-FJet-arxiv-current} (or just the equation of motion) that for small $\epsilon$ the dynamics for this model can be written as
\begin{equation}
\begin{bmatrix}
 du \\
 dv 
\end{bmatrix}
 = 
\begin{bmatrix}
 0 & 1 \\
 -\omega_0^2 & -2\gamma 
\end{bmatrix} 
\begin{bmatrix}
 u \\
 v 
\end{bmatrix}
\epsilon + {\cal O}(\epsilon^2) \, .
\label{eqn:app_eqnmotion}
\end{equation}
For this underdamped case, where $\gamma < \omega_0$, set $\omega = \sqrt{\omega_0^2 - \gamma^2}$.
The matrix in Eq.~\ref{eqn:app_eqnmotion} can be diagonalized using
\begin{equation*}
\begin{bmatrix}
\tilde{u} \\
\tilde{v} 
\end{bmatrix}
 = 
\frac{1}{2}
\begin{bmatrix}
 1 - i\gamma/\omega & -i/\omega \\
 1 + i\gamma/\omega & i/\omega 
\end{bmatrix} 
\begin{bmatrix}
 u \\
 v 
\end{bmatrix} \, ,
\end{equation*}
and similarly for $du$, $dv$.  
Ignoring ${\cal O}(\epsilon^2)$ terms, Eq.~\ref{eqn:app_eqnmotion} now becomes
\begin{equation}
\begin{bmatrix}
 d \tilde{u} \\
 d \tilde{v} 
\end{bmatrix}
 = 
\begin{bmatrix}
 \lambda_+ & 0 \\
 0 & \lambda_- 
\end{bmatrix} \\
\begin{bmatrix}
 \tilde{u} \\
 \tilde{v} 
\end{bmatrix}
\epsilon \, ,
\label{eqn:appA-dudv}
\end{equation}
where $\lambda_\pm = -\gamma \pm i\omega$.
The equations 
\begin{align*}
& \frac{ d \tilde{u} }{ \lambda_+ \tilde{u} } = \frac{ d \tilde{v} }{ \lambda_- \tilde{v} } 
\end{align*}
can now be directly integrated (see remarks in Sec.~\ref{sec:intfactors}), yielding the integration constant
\begin{align*}
r' 
& = (-\gamma - i\omega) \log \tilde{u} - (-\gamma + i\omega) \log \tilde{v} \, ,
\end{align*}
where ``log" is the natural logarithm.
It follows
\begin{align*}
r' & = - 2i \omega \log \sqrt{ 2\tilde{E} } + 2 \gamma \log e^{i\phi}  + \dots \, ,
\end{align*}
where dots indicate $u,v$-independent terms (which are ignored in this section), and $\phi$ and $\tilde{E}$ are defined by
\begin{align*}
\tilde{E} & = \frac{1}{2} [ (\omega u)^2 + (\gamma u + v)^2 ] \\
\tan \phi & = (\gamma u + v) / (\omega u) \, .
\end{align*}
The phase $\phi$ is restricted to a $2\pi$-interval, and so $\log e^{i\phi} = i (\phi - 2 n \pi)$ is defined using
Riemann sheets, with $n \in \mathbb{Z}$.
(An explicit minus sign is used so that as time increases and $\phi$ decreases, $n$ will increase.)

Throughout this paper, especially for the 2D oscillator examples, the integration constant will be a complex number,
with the real and imaginary parts each separately being constant.  Here, $r'$ is pure imaginary,
and the constant of interest is defined using the imaginary part of $r'$
\begin{align*}
r = -\frac{1}{\omega} {\Im} (r') \, .
\end{align*}
This directly leads to Eq.~\ref{eqn:r_under_main}.
Also, it is straightforward to verify that this is constant using the exact solution for $u(t)$.
In that case, as shown in the Supplement, $\tilde{E} \sim e^{-2\gamma t}$, $( \phi - 2 n \pi ) \sim -\omega t$,
and the time-dependence cancels out.


\section{Derivation of $r$ (overdamped)}
\label{sec:app_over}

\begin{figure}
\includegraphics[scale=0.30]{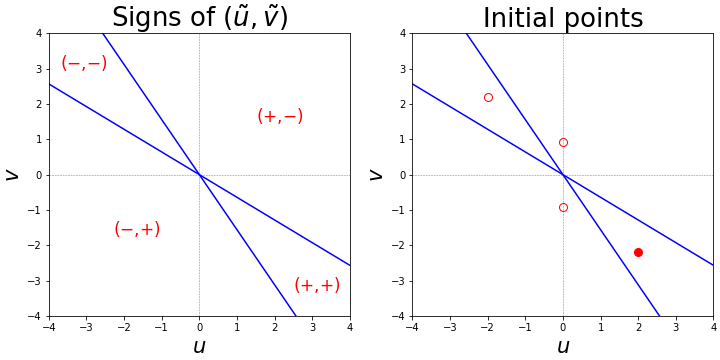}
\caption{ 
For the overdamped case, the $u,v$-plane is divided up into four regions, according to the signs of $(\tilde{u},\tilde{v})$.
In the left plot, for example, the region marked $+-$ corresponds to where $\tilde{u}>0$ and $\tilde{v}<0$.
These four regions are due to divergences in $r$, which occur where $\zeta u \pm (\gamma u   + v) = 0$.
The right plot displays an example of a starting point at $(\tilde{u}_0,\tilde{v}_0) = (1,1)$ (shown as a red dot),
and the other corresponding starting points (as open circles) at $(1,-1)$, $(-1,1)$, and $(-1,-1)$,
according to the discussion given.
}
\label{fig:regions}
\end{figure}

In this overdamped case, where $\gamma > \omega_0$, set $\zeta = \sqrt{ \gamma^2 - \omega_0^2 }$.
The matrix in Eq.~\ref{eqn:app_eqnmotion} can be diagonalized using
\begin{equation*}
\begin{bmatrix}
\tilde{u} \\
\tilde{v} 
\end{bmatrix}
 = 
\frac{1}{2 \zeta}
\begin{bmatrix}
 \gamma + \zeta & 1 \\
 -\gamma + \zeta & -1 
\end{bmatrix} 
\begin{bmatrix}
 u \\
 v 
\end{bmatrix} \, ,
\end{equation*}
and similarly for $du$, $dv$.  
Using these definitions for $\tilde{u},\tilde{v}$, and 
\begin{align*}
\lambda_\pm = -\gamma \pm \zeta \, ,
\end{align*}
Eq.~\ref{eqn:appA-dudv} can still be used.
In a similar manner to the underdamped case, the constant of motion can be computed from
\begin{align*}
r 
& = r_0 + \int^{\tilde{u}}_{\tilde{u}_0} \frac{ \lambda_- d \tilde{u}' }{ \tilde{u}' } - \int^{\tilde{v}}_{\tilde{v}_0} \frac{ \lambda_+ d \tilde{v}' }{ \tilde{v}' } \, .
\end{align*}
What is new in this case is that there are natural boundaries in the $u,v$ plane due to the zeros in $\tilde{u}$ and $\tilde{v}$, as a function of $u,v$;
these are illustrated in the left plot of Fig.~\ref{fig:regions}.  
Of course, during integration, these boundaries cannot be crossed, and so there are
technically four cases to consider when evaluating this integral (corresponding to the four regions shown in that plot).
However, by a judicious choice of an initial point, all four cases can be combined.
For example, if the initial point is $(u_0,v_0)$ in the +,+ region, then the initial point should be $(u_0,-v_0)$ in the $+-$ region, 
according to the nomenclature depicted in the right plot of Fig.~\ref{fig:regions}.
Proceeding in this manner, all four cases can be consolidated into
\begin{align*}
r 
 = & r_0 + \lambda_- \log \bigg\rvert \frac{\tilde{u}}{\tilde{u_0}} \bigg\rvert - \lambda_+ \log \bigg\rvert \frac{\tilde{v}}{\tilde{v_0}} \bigg\rvert \, .
\end{align*}
Next, letting the choice of $r_0$ cancel out the contribution from the initial point, this becomes
\begin{align}
r 
 = & \lambda_- \log | \tilde{u} | - \lambda_+ \log | \tilde{v} | \nonumber \\
 = & -( \zeta + \gamma ) \log | \zeta u + (\gamma u   + v) | \nonumber \\
    & - (\zeta - \gamma ) \log |  \zeta u - ( \gamma u + v) |  \, ,
\label{eqn:r_over}
\end{align}
where the constant $2 \zeta \log ( 2 \zeta )$ was ignored, and once again, ``log" is the natural logarithm, 

Also, if one is instead performing this integration numerically, it might not be convenient to coordinate the starting points for all four regions.  
As a result, there may be relative shifts of $r$ according to the region. 
However, that may be acceptable considering the regions are already separated by natural boundaries.

\section{Derivation of $r$ (critically damped)}
\label{sec:app_crit}

In the critically damped case, where $\gamma = \omega_0$, the general solution is
\begin{align}
u(t) & = (A + B t ) e^{-\gamma t} \, ,
\label{eqn:crit_exact}
\end{align}
with $A = u_0$ and $B = \gamma u_0 + v_0$ and $v = du/dt$.
Now notice that $\gamma u + v = B e^{-\gamma t}$, and that it
can be used along with the solution (Eq.~\ref{eqn:crit_exact}) to isolate the $t$-dependence as
\begin{align*}
& t = \left( \frac{u}{\gamma u + v} \right) - \left( \frac{u_0 }{\gamma u_0 + v_0 } \right) \, .
\end{align*}
Now use Eq.~\ref{eqn:r_alt1}, which in this case (i.e., $\gamma = \omega_0$) is
\begin{align*}
r' = \frac{1}{2} (\gamma u + v)^2 e^{2\gamma t} \, .
\end{align*}
Upon setting $r'' = (\log (2r'))/2$, substituting in for $t$, and ignoring $u,v$-independent shifts, the constant for this case is found to be
\begin{align*}
r'' & = \log | \gamma u + v | + \left( \frac{ \gamma u }{\gamma u + v} \right) \, .
\end{align*}

\section{Alternative Solutions}
\label{sec:alt_solutions}

With the wisdom afforded by an existing solution for the 1D damped harmonic oscillator, it becomes possible to more easily discover new ones.
Indeed, in the expressions for $r$ (cf. Eqs.~\ref{eqn:r_under_main},\ref{eqn:r_over_main},\ref{eqn:r_critical_main}), the variable combination $\gamma u + v$ 
appeared.  
Thus, the variable $w = \gamma u + v$ is introduced, and leads to the equations of motion
\begin{align*}
\frac{d}{dt}
\begin{bmatrix}
u \\
w 
\end{bmatrix}
& = 
\begin{bmatrix}
 - \gamma & 1 \\
 (\gamma^2 - \omega_0^2)  & - \gamma 
\end{bmatrix} 
\begin{bmatrix}
u \\
w 
\end{bmatrix} \, .
\end{align*}
The diagonal terms can be removed by setting $u  = \tilde{u} e^{-\gamma t}$ and $w  = \tilde{w} e^{-\gamma t}$
\begin{align}
\frac{d}{dt}
\begin{bmatrix}
\tilde{u} \\
\tilde{w} 
\end{bmatrix}
& = 
\begin{bmatrix}
 0  & 1  \\
 (\gamma^2 - \omega_0^2)  &  0
\end{bmatrix} 
\begin{bmatrix}
\tilde{u} \\
\tilde{w} 
\end{bmatrix} \, .
\label{eqn:alternate-main}
\end{align}
At this point there are now at least two approaches that can be used to solve for a constant of motion, and are given next.

\subsection{Alternative \#1}
\label{sec:alt_solutions_1}

The first alternative approach starts from Eq.~\ref{eqn:alternate-main} and seeks to integrate
\begin{align*}
\frac{ d\tilde{u} }{ \tilde{w} } = \frac{ d\tilde{w} }{ (\gamma^2 - \omega_0^2) \tilde{u} } \, .
\end{align*}
This can be directly integrated without need for an integrating factor (see discussion in Sec.~\ref{sec:intfactors}), yielding the constant
\begin{align}
r' 
& = \frac{1}{2} \left[ ( \omega_0^2 - \gamma^2 ) u^2 + (\gamma u + v)^2 \right] e^{2\gamma t} \, .
\label{eqn:r_alt1}
\end{align}
The reader should notice that no assumptions were made about it being either the underdamped, overdamped, or critically damped case.

In a comparison to the constants already derived in Secs.~\ref{sec:app_under},\ref{sec:app_over}, and \ref{sec:app_crit}, the obvious differences are that there
is an explicit $t$-dependence, and also that it should be transformed as $r = \log (2r')$.
The $t$-dependence can be removed using an exact solution for any of the three damping cases.
Actually, this was the approach used for the critically damped case in App.~\ref{sec:app_crit}.

\subsection{Alternative \#2 (underdamped)}
\label{sec:alt_solutions_2}

Another way to exploit Eq.~\ref{eqn:alternate-main} is to use a specific solution for $u$, $w$.
This is demonstrated here for the underdamped case
\begin{align*}
\frac{d}{dt}
\begin{bmatrix}
\tilde{u} \\
\tilde{w} 
\end{bmatrix}
& = 
\begin{bmatrix}
 0  & 1  \\
 -\omega^2  &  0
\end{bmatrix} 
\begin{bmatrix}
\tilde{u} \\
\tilde{w} 
\end{bmatrix} \, .
\end{align*}
Setting $\tilde{u}' = \omega \tilde{u}$ and $t' = \omega t$, yields
\begin{align*}
\frac{d}{dt'}
\begin{bmatrix}
\tilde{u}' \\
\tilde{w} 
\end{bmatrix}
& = 
\begin{bmatrix}
 0  & 1  \\
 -1  &  0
\end{bmatrix} 
\begin{bmatrix}
\tilde{u}' \\
\tilde{w} 
\end{bmatrix} \, .
\end{align*}
Writing the solution as (with parameters $A,\beta$)
\begin{align*}
\tilde{u}' & = A\cos( t' - \beta) \\
\tilde{w} & = -A\sin( t' - \beta) \, ,
\end{align*}
suggests the complex variables
\begin{align*}
z          & = \tilde{u}' + i \tilde{w} = A e^{ -i(t' - \beta) }  \\
\bar{z} & = \tilde{u}' - i \tilde{w} = A e^{ i(t' - \beta) }  \, .
\end{align*}
The explicit $t'$-dependence can now be removed by forming
\begin{align*}
z \bar{z} & = A^2 \\
& = \tilde{u}'^2 + \tilde{w}^2 \\
& = \left[ \omega^2 u^2 + (\gamma u + v)^2 \right] e^{2\gamma t} \, .
\end{align*}
In this case the constant $A^2$ supplants the role of $r'$.
For this underdamped case, the $t$-dependence can be removed using
\begin{align*}
\tan ( \omega t - \beta ) 
& = -\tilde{w} / \tilde{u}'  = - (\gamma u + v) / ( \omega u ) \, ,
\end{align*}
and then taking into account the same precautions as before owing to periodicity.


\subsection{Alternative \#2 (overdamped)}
\label{sec:alt_solutions_2b}

In a similar manner to that of the preceeding subsection, a demonstration can also be made for the overdamped case, where $\gamma > \omega_0$.
The equation of motion becomes, with $\zeta = \sqrt{\gamma^2 - \omega_0^2}$, 
\begin{align*}
\frac{d}{dt}
\begin{bmatrix}
\tilde{u} \\
\tilde{w} 
\end{bmatrix}
& = 
\begin{bmatrix}
 0  & 1  \\
 \zeta^2  &  0
\end{bmatrix} 
\begin{bmatrix}
\tilde{u} \\
\tilde{w} 
\end{bmatrix} \, .
\end{align*}
An explicit solution may be written as
\begin{align*}
\tilde{u} & = A' e^{\zeta t} + B' e^{ -\zeta t} \, ,
\end{align*}
for constants $A'$ and $B'$, from which follows
\begin{align*}
\zeta u - w & = 2 \zeta B' e^{ -\gamma t -\zeta t} \\
\zeta u + w & = 2 \zeta A' e^{-\gamma t + \zeta t}  \, .
\end{align*}
By taking the log of each equation and multiplying the top by $(\gamma - \zeta)$ and the bottom
by $(\gamma + \zeta)$, it becomes possible to eliminate the time dependence and isolate a constant as
\begin{align*}
r' & = (\gamma - \zeta) \log | \zeta u - w | - (\gamma + \zeta) \log | \zeta u + w| \, ,
\end{align*}
which is in agreement with that found earlier (cf. Eq.~\ref{eqn:r_over}).



\bibliography{../../../../../../Biblio/BIB_proj4_FJet,../../../../../../Biblio/BIB_MZ,../../../../../../Biblio/BIB_ML,../../../../../../Biblio/BIB_math_Lie}

\end{document}